\documentclass{article}

\PassOptionsToPackage{numbers,sort&compress}{natbib}
\usepackage[preprint]{neurips_2026}

\usepackage[utf8]{inputenc} % allow utf-8 input
\usepackage[T1]{fontenc}    % use 8-bit T1 fonts
\usepackage{mathtools}      % amsmath + \coloneqq
\usepackage{amsmath}
\usepackage{amssymb}
\usepackage{hyperref}       % hyperlinks
\usepackage{url}            % simple URL typesetting
\usepackage{booktabs}       % professional-quality tables
\usepackage{amsfonts}       % blackboard math symbols
\usepackage{nicefrac}       % compact symbols for 1/2, etc.
\usepackage{microtype}      % microtypography
\usepackage{xcolor}         % colors
\usepackage{float}

\usepackage{graphicx}
\usepackage{subcaption}
\usepackage{tikz}
\usetikzlibrary{positioning, arrows.meta}
\title{Grammar Sparsity Sets How Transformers Learn Random Languages}

\author{%
  Edgar Andres Hernandez Moreno \\
  Universitat de Barcelona \\
  \texttt{ehernamo48@alumnes.ub.edu} \\
  \And
  Francesco Cagnetta \\
  Simplex \\
}

\begin{document}

\maketitle

\begin{abstract}
We study language acquisition by transformers trained on the Random Language Model: a synthetic hierarchical grammar with random production rules, whose sparsity is controlled by a single parameter and whose conditional entropies are known exactly. On depth-two grammars, we find that sparsity sets both the order and the speed of learning. Dense grammars are learnt one level at a time: a token is predicted from its sibling after $t\propto V$ steps and from the full context only after $t\propto V^{3}$, so the two stages are separated by a plateau that widens with $V$. Sparse grammars are learnt all at once, at $t\propto\sqrt V$, with a smaller exponent than even the first stage of a dense one. The exponents $1$ and $1/2$ already appear at depth one, where we derive them for a linear model trained with Adam.
\end{abstract}
\begin{figure*}[h!]
  \centering
  \begin{subfigure}[t]{0.3\textwidth}
    \vspace{15pt}% [t] baseline at the top (+15pt to align the tree with the plot area)
    \centering
    % \documentclass{standalone}
% \usepackage{tikz}
% \usetikzlibrary{positioning, arrows.meta}
% \begin{document}
\begin{tikzpicture}[
    baseline=(3-0.base),
    varnode/.style={circle, draw=black, thick, minimum size=8.5mm, inner sep=0pt, font=\small},
    hidnode/.style={varnode, minimum size=10mm, fill=black!12},
    fac/.style={rectangle, draw, thick, fill=black, minimum size=2mm},
    plain/.style={line width=0.25mm}
]

\tikzset{
  mytriangle/.tip = {Triangle[angle=60:2mm]},
  arrowdown/.style = {->, >={mytriangle}, line width=0.25mm}
}

% Labels: #1 = level (1 = root, \layers = leaves), #2 = node index (0-based)
% root -> R, hidden level-1 symbols -> H_i, observed leaves -> X_i
\providecommand{\getlabel}{}
\renewcommand{\getlabel}[2]{%
    \pgfmathtruncatemacro\labidx{#2+1}%
    \ifnum#1=1 $R$%
    \else\ifnum#1=\layers $X_{\labidx}$%
    \else $Z_{\labidx}$%
    \fi\fi
}

% Layout settings (same as rhm_tree.tex, so the subtree keeps its size there)
\providecommand\layers{}\renewcommand\layers{3}
\providecommand\initialdistance{}\renewcommand\initialdistance{.25cm}
\providecommand\laySpace{}\renewcommand\laySpace{.06cm}

%%% Create nodes (only the subtree rooted at X^{(1)}_1, i.e. node 0 of level 1)
\foreach \layer in {\layers,...,1} {
    \pgfmathsetmacro\offX{\initialdistance/(2^(\layer+1))};
    \pgfmathsetmacro\Xdistance{\initialdistance/(2^\layer)};
    \pgfmathsetmacro\numNodes{2^(\layer-1)-1};
    \foreach \n in {0,...,\numNodes} {
        \pgfmathsetmacro\Xposition{\offX + \Xdistance * \n};
        \pgfmathsetmacro\Yposition{-\laySpace * \layer};
        \ifnum\layer<\layers
            \node [hidnode] (\layer-\n) at (\Xposition, \Yposition) {\getlabel{\layer}{\n}};
        \else
            \node [varnode] (\layer-\n) at (\Xposition, \Yposition) {\getlabel{\layer}{\n}};
        \fi
        \ifnum\layer<\layers
            \node[fac] (f\layer-\n) at (\Xposition, \Yposition-0.8) {};
        \fi
    }
}

%%% Draw connections
\foreach \layer in {\layers,...,1} {
    \pgfmathsetmacro\numNodes{2^(\layer-1)-1};
    \foreach \n in {0,...,\numNodes} {
        \pgfmathtruncatemacro\prevLay{\layer-1};
        \pgfmathtruncatemacro\uppNode{int(\n/2)};
        \ifnum\layer>1
            \draw[arrowdown] (f\prevLay-\uppNode) -- (\layer-\n);
        \fi
        \ifnum\layer<\layers
            \draw[plain] (\layer-\n) -- (f\layer-\n);
        \fi
    }
}

\end{tikzpicture}
% \end{document}
    \label{fig:main-tree}
  \end{subfigure}
  \begin{subfigure}[t]{0.6\textwidth}
    \vspace{0pt}%
    \centering
    \includegraphics[width=\linewidth]{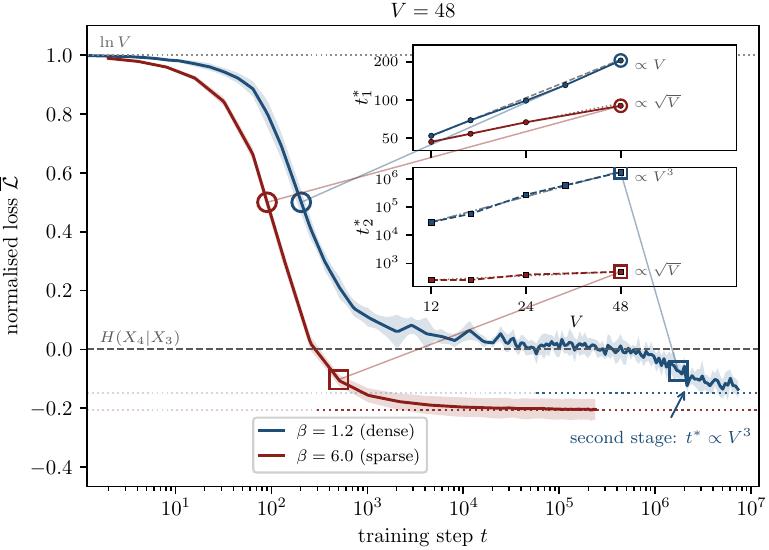}
    \label{fig:main-learning}
  \end{subfigure}
  \caption{\label{fig:main}\textbf{The sparsity of the grammar sets both the schedule and
the time complexity.}
\textbf{Left: The RLM.} A root symbol $R$ is expanded recursively through the production
rules $\mathcal Q_a$ of Eq.~\eqref{eq:rlm-transition} (black squares); first into a pair of latent variables $(Z_1, Z_2)$, then into the observed sequence
$(X_1,\dots,X_4)$.
\textbf{Right: Learning dynamics of a deep transformer for a dense ($\beta=1.2$, $V=48$, blue) and a sparse ($\beta=6.0$, $V=48$, red) grammar.}
The test loss is normalised, $\bar{\mathcal L}=(\mathcal L-H_1)/(\ln V-H_1)$: it equals $1$ for the uniform predictor and $0$ at the sibling entropy
$H_1=H(X_4\mid X_3)$ (dashed). Dotted lines mark the full-context floor
$H_2=H(X_4\mid X_1,X_2,X_3)$ of each grammar. The dense grammar descends in two stages
separated by a plateau at $H_1$, the sparse one in a single step. Markers locate the two
time complexities, $t_1^*$ (circles) and $t_2^*$ (squares), shown against $V$ in the insets.
Averages over $8$ realisations of the grammar and transformer initialisation; bands give one standard deviation.}
  \label{fig:main-gamma}
\end{figure*}

\section{Introduction}

Sentences are formed hierarchically, words into phrases and phrases into clauses~\cite{chomsky1957,everaert2015,ding2016}, and whether this structure can be learned from the statistics of sentences alone has long been debated~\cite{chomsky1980,saffran1996,saffran2018,futrell2025,futrell2026}. Large language models have shown that it can: trained on next-token prediction alone, they develop representations that track the syntactic tree of the sentence~\cite{tenney2019,manning2020} and implement parsing-like computations~\cite{zhao2023,allenzhu2023}. How these representations are acquired during training is far less clear: which statistics of the text drive their emergence, and how much data or training time each of them costs.
Since the structure of natural text cannot be varied, a theory of language acquisition calls for synthetic data with controlled hierarchical structure~\cite{mossel2016,malach2018,cagnetta2024rhm}, where the progress of a learner can be measured precisely~\cite{cagnetta2024language,parley2026,korchinski2026}.

Here we consider the Random Language Model (RLM)~\cite{degiuli2019prl,degiuli2019jpa, giorlandino2026}: a grammar-like data generating process, where sequences are generated by expanding a tree of latent variables via a prescribed set of rules.
The structure of RLM data is controlled by a sparsity parameter $\beta$: As $\beta$ increases, the grammar goes from \emph{dense}, where the effective number of rules per symbol grows as a power of the vocabulary size $V$, to \emph{sparse}, where each symbol uses $O(1)$ rules.
We ask how the grammar sparsity shapes the learning dynamics of transformers: in which order the levels of the hierarchy are acquired, and how the training time of each level scales with the vocabulary size.

\vspace{-5pt}
\paragraph{Our Contributions.}
We train deep transformers on $4$-token sequences generated by a depth-$2$ RLM to predict the last token from the preceding context.
We measure the training times at which the test loss reaches the entropy of the last token given its sibling, $H(X_4|X_3)$, then its entropy given the full context, $H(X_4|X_1,X_2,X_3)$, as functions of the vocabulary size $V$ (Fig.~\ref{fig:main}).
We find that the sparsity of the rules sets both the schedule and the speed of learning:
\begin{itemize}
    \item \textbf{Dense grammars are learnt in stages,} one level of the hierarchy at a time. The sibling statistics are acquired at $t_1^*\propto V$ (Fig.~\ref{fig:main}, top inset), whereas the information carried by the first two tokens is acquired at $t_2^*\propto V^{3}$ (bottom inset).
    \item \textbf{Sparse grammars are learnt in a single step,} at $t^*\propto\sqrt V$ (Fig.~\ref{fig:main}, insets).
    \item \textbf{The different regimes are already observed at depth one:} Even when trained to predict the second token of a depth-$1$ RLM from the first, transformers still take $t^*\propto V$ on dense grammars and $t^*\propto\sqrt V$ on sparse ones (Fig.~\ref{fig:L1}). We derive both exponents analytically for a linear model trained with Adam (Sec.~\ref{sec:L1}, App.~\ref{app:adam}).
\end{itemize}

\section{Random Language Model (RLM)}
\label{sec:rlm}

\textbf{Data-Generating Process.}
We use a layered variant of the Random Language Model~\cite{degiuli2019prl,degiuli2019jpa,giorlandino2026} (RLM) to generate the leaves of a  depth-$L$ binary tree iteratively (Fig.~\ref{fig:main}, left): the root $R$ expands into two latent variables $(Z_1,Z_2)$, each independently generating its two children. The expansion follows production rules $a\to bc$, where $a$ is the parent symbol and $bc$ is an ordered pair of children, all taken from the same vocabulary of size $V$. Given the parent $a$, the distribution $\mathcal Q_a$ over child pairs is defined as
\begin{equation}\label{eq:rlm-transition}
\mathcal Q_a(b,c)\coloneqq\frac{e^{\beta\mathcal E_{a,b,c}}}{\sum_{b',c'=1}^V e^{\beta\mathcal E_{a,b',c'}}},\qquad \mathcal E_{a,b,c}\overset{\text{i.i.d.}}{\sim}\mathcal N(0,\ln V),
\end{equation}
where the random couplings $\mathcal E$ are drawn once and then kept fixed.

\begin{figure}[h!]
\centering
\includegraphics[width=\textwidth]{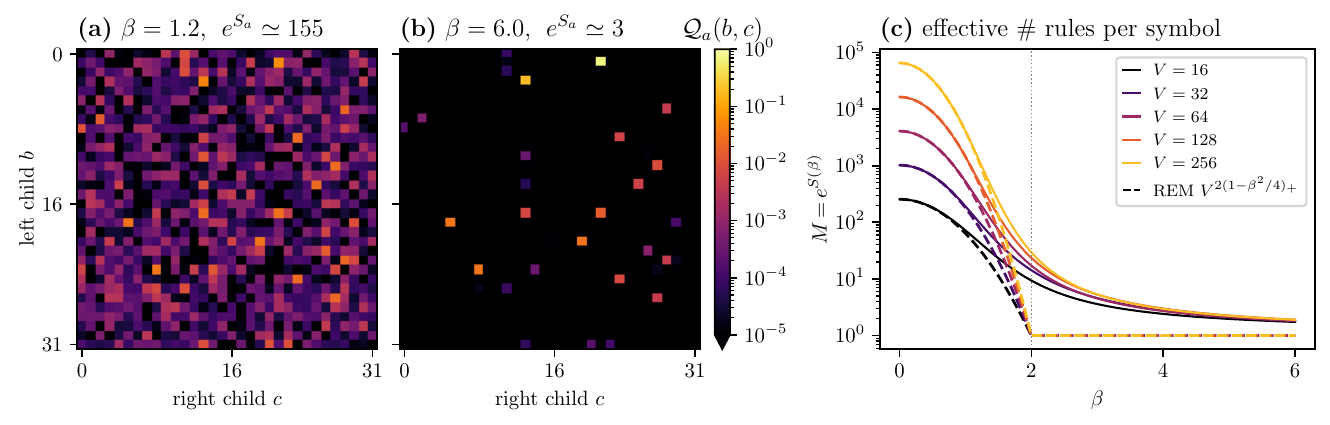}
\caption{\label{fig:rules}\textbf{A single parameter controls the sparsity of the grammar.}
\textbf{(a,\,b)} The rule matrix $\mathcal Q_a(b,c)$ of one parent symbol at $V=32$ and $\beta=1.2,6.0$ on a shared logarithmic colour scale. At $\beta=1.2$, many pairs carry appreciable weight, so the grammar is
\emph{dense}; at $\beta=6$ the weight condenses onto a handful of pairs and the grammar is \emph{sparse}.
\textbf{(c)} The effective number of rules per symbol, $M(\beta)$ of
Eq.~\eqref{eq:rule-entropy}, averaged over parent symbols and over realisations of the
couplings. Dashed lines give the $V\to\infty$ prediction of Eq.~\eqref{eq:rem}, with the freezing transition at $\beta=2$~\cite{giorlandino2026} marked by a vertical line.}
\end{figure}

\textbf{Sparsity.}
$\beta$ controls how many rules carry appreciable probability, quantified by the entropy of $\mathcal{Q}_a$,
\begin{equation}\label{eq:rule-entropy}
S_a(\beta)\coloneqq-\sum_{b,c=1}^V\mathcal Q_a(b,c)\ln\mathcal Q_a(b,c),\qquad S(\beta)\coloneqq\frac1V\sum_{a=1}^V S_a(\beta),
\end{equation}
so that $M(\beta)\coloneqq e^{S(\beta)}$ is the effective number of rules per symbol (Fig.~\ref{fig:rules}).
$S_a$ decreases monotonically from $S_a(0)=2\ln V$ (all $V^2$ rules equiprobable) to $S_a(\infty)=0$ (a single rule per symbol).
By a mapping identified in~\cite{giorlandino2026}, each $\mathcal Q_a$ is the Gibbs measure of a random energy model with $V^2$ levels~\cite{derrida1981}. Hence, as $V\to\infty$,
\begin{align}\label{eq:rem}
S_a(\beta)/(2\ln V)\to \begin{cases}1-\beta^2/4& \beta\le2, \\ 0&\beta>2.\end{cases} \Rightarrow M(\beta) = \begin{cases} O\left(V^{2-\beta^2/2}\right)& \beta\le2, \\ O\left( 1 \right)&\beta>2.\end{cases}
\end{align}

\section{Learning schedule and time scales at $L=2$}
\label{sec:L2}

\textbf{Task.}
We fix $L=2$, so that the observed sequence $(X_1,\dots,X_4)$ descends from the root $R$ through the two latent symbols $Z_1$ and $Z_2$, parents of $(X_1,X_2)$ and $(X_3,X_4)$ respectively (Fig.~\ref{fig:main}, left).
The task is to predict the last token $X_4$ from the context $(X_1,X_2,X_3)$.
A two-layer decoder-only transformer $\hat p_\theta(x_4\mid x_1,x_2,x_3)$ is trained with Adam on the cross-entropy, on a fresh batch of sequences at every step, so that the training step $t$ also counts, up to the batch size, the samples seen; all hyperparameters are held fixed as $V$ and $\beta$ vary (App.~\ref{app:protocol}).
We monitor the test loss
\begin{equation}\label{eq:test-loss}
\mathcal L(\theta)\coloneqq-\mathbb E\big[\ln\hat p_\theta(X_4\mid X_1,X_2,X_3)\big]\;\ge\;H_2\coloneqq H(X_4\mid X_1,X_2,X_3),
\end{equation}
with equality if and only if $\hat p_\theta$ coincides with the true conditional almost surely.
Two further reference values organise the learning curves: $\ln V$, the loss of the uniform predictor, from which the network starts, and $H_1\coloneqq H(X_4\mid X_3)\ge H_2$, the best loss achievable from the sibling $X_3$ alone. The two conditional distributions are qualitatively different: $X_3$ and $X_4$ share the parent $Z_2$, so their dependence is that of a single rule, whereas the cousins $(X_1,X_2)$ inform $X_4$ only through $Z_1$, and the gap $H_1-H_2$ measures the information carried by the latent variable. Both entropies are computed exactly from the rules of each realisation (App.~\ref{app:protocol}).

\textbf{Two learning schedules.}
Fig.~\ref{fig:main} (right) shows the test loss at $V=48$ on a dense ($\beta=1.2$) and on a sparse ($\beta=6$) grammar, in the normalised form $\bar{\mathcal L}\coloneqq(\mathcal L-H_1)/(\ln V-H_1)$.
The dense grammar is learnt in two stages, mirroring the phenomenology observed in~\cite{cagnetta2024language}: 
First, the loss descends to $H_1$, where the network predicts $X_4$ from $X_3$ as well as possible while still ignoring $(X_1,X_2)$;
Then, after roughly three decades of training time on this plateau, it descends to $H_2$, when the cousins are finally exploited.
The order follows the tree: the level-$1$ rule linking $X_3$ to $X_4$ is acquired before the root-level rule linking the two halves of the sequence.
The sparse grammar shows no plateau: the loss drops from $\ln V$ to $H_2$ in a single step, i.e., sibling and cousins are exploited together.
This is not because the second stage is less informative in the sparse grammar: relative to the first stage, the gap $H_1-H_2$ is larger at $\beta=6$ than at $\beta=1.2$ (compare the floors in Fig.~\ref{fig:main}).

\textbf{Time complexities.}
To quantify the speed of learning, we define $t_1^*$ as the training step at which $\bar{\mathcal L}$ first crosses $1/2$, halfway between $\ln V$ and $H_1$ (circles in Fig.~\ref{fig:main}), and $t_2^*$ as the step at which $\mathcal L$ first crosses the halfway point between $H_1$ and $H_2$ (squares); both definitions apply whether or not a plateau is present.
The insets show $t_1^*$ (top) and $t_2^*$ (bottom) against $V$.
On the dense grammar, $t_1^*\propto V$ and $t_2^*\propto V^{3}$: each stage costs a power of the vocabulary size, and since the two powers differ, the plateau widens as $V$ grows.
On the sparse grammar, both times scale as $\sqrt V$ and differ only by an $O(1)$ factor: the whole grammar is acquired at once, with a smaller exponent than even the first stage of the dense grammar.
App.~\ref{app:numerics} shows the collapse of the learning curves for all $V$ against $t/V^{s}$, with fitted exponents $s_1=1.00\pm0.02$ and $s_2=3.04\pm0.34$ at $\beta=1.2$ and $s=0.44\pm0.03$ at $\beta=6$, and the intermediate case $\beta=2$, where the two stages are still distinct but separated by a smaller power of $V$.

\section{Time complexity scaling at $L=1$}
\label{sec:L1}

\begin{figure}[h!]
  \centering
  \includegraphics[width=\textwidth]{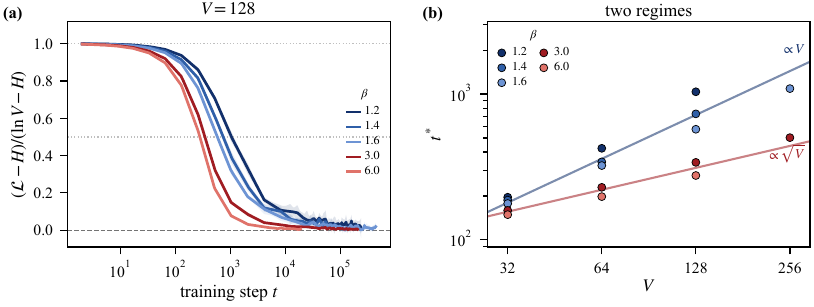}
  \caption{\label{fig:L1}\textbf{Both regimes are present at $L=1$.} Transformers trained on a depth-$1$ RLM to predict $X_2$ from $X_1$.
  \textbf{(a)} Normalised test loss $(\mathcal L-H)/(\ln V-H)$ at $V=128$, with
  $H=H(X_2\mid X_1)$ computed exactly from the rules of each realisation, for three values
  of $\beta$ in the dense phase ($1.2$, $1.4$, $1.6$; blues) and two in the sparse phase
  ($3$, $6$; reds). Dotted lines mark the levels $1$ and $1/2$.
  \textbf{(b)} The crossing time $t^*$ against $V$ for $V=32$ to $256$. Solid lines show the two
  scalings $t^*\propto V$ and $t^*\propto\sqrt V$ derived for a linear model trained with
  Adam in App.~\ref{app:adam}; within a phase, different $\beta$ shift $t^*$ by an $O(1)$
  factor without changing the exponent. Averages over repetitions, bands giving one
  standard deviation; protocol in App.~\ref{app:protocol}.}
\end{figure}

The two exponents of Fig.~\ref{fig:main} are not a consequence of the hierarchy.
Fig.~\ref{fig:L1} shows transformers trained on a depth-$1$ RLM to predict the second token from the first: the loss descends in one step at every $\beta$, from $\ln V$ to $H\coloneqq H(X_2\mid X_1)$ (panel a), yet the time $t^*$ at which the normalised loss reaches one half still separates the two phases (panel b), growing as $V$ in the dense phase and as $\sqrt V$ in the sparse one---the exponents of the first stage and of the single step at depth two.

Both exponents follow from a linear model, a table of logits $\theta_{ab}$ for predicting $b$ after $a$, trained with Adam (batch size $B$, learning rate $\eta$) on the cross-entropy (see App.~\ref{app:adam} for details).
Replacing Adam's running moments by their population values, a coordinate whose gradient has mean $\mu_{ab}$ and standard deviation $\sigma_{ab}$ moves by $\eta\,\mu_{ab}\sqrt B/\sigma_{ab}$ per step as long as this signal-to-noise ratio is small. Hence, covering the logit distance $D_{ab}$ to its target takes $t^*\asymp D_{ab}\,\sigma_{ab}/(\eta\,|\mu_{ab}|\sqrt B)$ steps.
In the sparse phase, a logit receives a gradient only from the fraction $1/V$ of samples with the right first token, which dilutes the signal by $1/V$ but the noise only by $1/\sqrt V$, hence $t^*\propto\sqrt V$ up to a logarithm. In the dense phase, signal and distance to travel are both proportional to the weak correlation between the two tokens, which cancels, and the noise, set by the joint probability, leaves $t^*\propto V$ at every $\beta$ in the phase. Plain SGD, which does not divide by the noise, would give $V$ and $V^2$ instead (App.~\ref{app:adam}).

\section{Conclusions}
\label{sec:conclusions}

We found that, for transformers trained on hierarchical grammars with controlled sparsity, the latter sets both the order and the speed at which the latent structure of the data is acquired. Dense grammars are learnt level by level, each level at a higher power of $V$, so the plateau between stages widens with $V$; sparse grammars are learnt in a single step, with a smaller exponent.
Two directions remain open. The first is the role of depth in the acquisition: the preliminary evidence of App.~\ref{app:depth} suggests that deep in the sparse phase a single attention layer suffices, whereas close to the transition a second layer is needed to reach the full-context floor. The second is the mechanism by which the latent variables are learnt, which sets the exponent of the second stage: our argument covers only the first stage, and it remains to be seen whether the second follows the correlation-based mechanism of~\cite{cagnetta2024language}, in which each level of the hierarchy is acquired once the correlations it induces between tokens are resolved from the sampling noise.

\begin{ack}
This project started during EAHM's research internship at SISSA, as part of EAHM's B.S. in Engineering Physics at the Instituto Tecnol\'ogico y de Estudios Superiores de Monterrey (Tecnol\'ogico de Monterrey), during which most of the work was carried out. The authors would like to thank Sebastian Goldt for useful discussions. For part of this project, the work of FC was supported by the European Union's Horizon Europe program under the Marie Sk\l{}odowska-Curie grant agreement No.~101154584.
\end{ack}

\bibliographystyle{unsrt}
\bibliography{references}

%%%%%%%%%%%%%%%%%%%%%%%%%%%%%%%%%%%%%%%%%%%%%%%%%%%%%%%%%%%%

\appendix

\section{Details of the simulations}
\label{app:protocol}

Code and configuration files reproducing every figure of this paper will be
made publicly available upon acceptance.

\paragraph{Data.}
Sequences are sampled online from the depth-$L$ RLM of Sec.~\ref{sec:rlm}: a
fresh batch is drawn from the grammar at every training step, so there is no
finite training set and no epochs. The root is uniform over the vocabulary and
a single realisation of the couplings, Eq.~\eqref{eq:rlm-transition}, defines
one tensor of production rules shared by all levels of the tree. Each
repetition uses an independent grammar, an independent weight initialisation
and an independent data stream. we report averages over $8$ repetitions, with shaded bands giving
one standard deviation across repetitions.

\paragraph{Architecture.}
We use a decoder-only causal transformer. Tokens and absolute positions are
embedded with learned embeddings of dimension $d$, initialised as
$\mathcal N(0,0.02^2)$ and summed. The trunk is a stack of pre-LayerNorm
residual blocks, each combining causal multi-head self-attention with $4$ heads
(bias-free key, query, value and output projections) and a two-layer perceptron
of width $4d$ with a single ReLU and no biases, in the NTK parametrisation. A
final LayerNorm is followed by a linear head onto the $V$ logits, with bias and
untied from the embedding, so that the loss at initialisation is close to
$\ln V$. Dropout is disabled. Only the logits at the last input position enter
the loss: for $L=2$ the input is the context $(X_1,X_2,X_3)$ and the target is
$X_4$; for $L=1$ the input is $X_1$ and the target is $X_2$.

\paragraph{Training.}
We minimise the cross-entropy of the last token with Adam (default
$\beta_1=0.9$, $\beta_2=0.999$, $\epsilon=10^{-8}$, no weight decay) at a
constant learning rate of $10^{-4}$, without warmup or decay. The test loss is
the mean cross-entropy on freshly sampled sequences of the same grammar,
evaluated at logarithmically spaced steps early in training and at regular
intervals afterwards. The number of steps is chosen so that every run reaches
its floor. All hyperparameters are held fixed as $V$ and $\beta$ are varied.

Three settings are used, and curves from different settings are never placed on
the same axes:
\begin{center}\small
\begin{tabular}{lccccc}
\toprule
 & blocks & $d$ & batch & test samples & used in \\
\midrule
main text ($\beta=1.2$, $6.0$) & $2$ & $256$ & $128$ & $4096$ & Figs.~\ref{fig:main}, \ref{fig:app-b12-s1}, \ref{fig:app-b12-s2}, \ref{fig:app-b60}, \ref{fig:app-depth-b60} \\
smaller model ($\beta=2.0$)    & $2$ & $128$ & $\phantom{0}64$ & $\phantom{0}512$ & Figs.~\ref{fig:app-b20-s1}, \ref{fig:app-b20-s2}, \ref{fig:app-depth-b20} \\
depth-one task ($L=1$)         & $1$ & $128$ & $128$ & $4096$ & Fig.~\ref{fig:L1} \\
\bottomrule
\end{tabular}
\end{center}
The first has $1.60\times10^{6}$ parameters at $V=48$, the second
$4.03\times10^{5}$ at $V=32$. Runs at $\beta=2$ exist only in the second
setting. In App.~\ref{app:depth} the number of blocks is varied, one against
two, at otherwise identical settings: the paired runs at $\beta=2$ use the
smaller model, those at $\beta=6$ the main one.

\paragraph{Entropy floors and normalised loss.}
The reference entropies are computed exactly from the rule tensor of each
repetition by summing over the latent variables, not estimated from samples.
For $L=2$ these are $H_1=H(X_4|X_3)$ and $H_2=H(X_4|X_1,X_2,X_3)$; for $L=1$,
$H=H(X_2|X_1)$. The normalised loss of the first stage is
$(\mathcal L-H_1)/(\ln V-H_1)$, that of the second stage
$(\mathcal L-H_2)/(H_1-H_2)$, and that of the single stage in the sparse regime
and at $L=1$, $(\mathcal L-H_2)/(\ln V-H_2)$ and $(\mathcal L-H)/(\ln V-H)$
respectively. Each runs from $1$ at the start of the stage to $0$ at its end.

\paragraph{Time scales.}
Repetitions are interpolated onto a common logarithmic grid in $t$ and
averaged; $t^*_\ell$ is the first downward crossing of $1/2$ of the averaged
normalised curve, obtained by log-linear interpolation between measurements.
The exponents $s_\ell$ in $t^*_\ell\propto V^{s_\ell}$ are least-squares fits of
$\ln t^*_\ell$ against $\ln V$, and the quoted uncertainties are standard
deviations over $1000$ bootstrap resamples of the repetitions, the averaging
and the fit being repeated for each resample. The exponents depend somewhat on
the crossing level used to define $t^*$: at $\beta=1.2$ the first-stage
exponent ranges from $1.05$ at the half-way point to $1.45$ at the $10\%$
level, a sign that the curves do not collapse perfectly. The half-way point is
used throughout.

\section{Details of the numerical results}
\label{app:numerics}

This appendix reports the learning curves behind the time scales of
Fig.~\ref{fig:main}, resolved by vocabulary size, together with the intermediate
grammar $\beta=2$; there is one figure for each stage of each grammar. Every
figure shows the same data twice: on the left the normalised loss against the
training step, on the right against the rescaled time $t/V^{s}$, with $s$ the
fitted exponent. The right panel collapses onto a single master curve precisely
when $t^{*}\propto V^{s}$.

The normalisation of the vertical axis is the one defined in
App.~\ref{app:protocol}. For a stage that starts at a loss $\mathcal L_{\rm i}$
and ends at a loss $\mathcal L_{\rm f}$, we plot
$(\mathcal L-\mathcal L_{\rm f})/(\mathcal L_{\rm i}-\mathcal L_{\rm f})$, which
runs from $1$ at the beginning of the stage to $0$ at its end and removes the
dependence of the entropies on $V$ and $\beta$. Concretely, the first stage
starts at the uniform predictor and ends at the sibling entropy, so
$\mathcal L_{\rm i}=\ln V$ and $\mathcal L_{\rm f}=H_1$; the second stage starts
at the sibling entropy and ends at the full-context entropy, so
$\mathcal L_{\rm i}=H_1$ and $\mathcal L_{\rm f}=H_2$; and the single stage of
the sparse regime runs from the uniform predictor directly to the full-context
entropy, $\mathcal L_{\rm i}=\ln V$ and $\mathcal L_{\rm f}=H_2$. Both
entropies are computed exactly from the rules of each realisation. Curves are
averages over realisations on a common logarithmic grid in $t$, and shaded bands
give one standard deviation across realisations.

\begin{figure}[h!]\centering
\includegraphics[width=\textwidth]{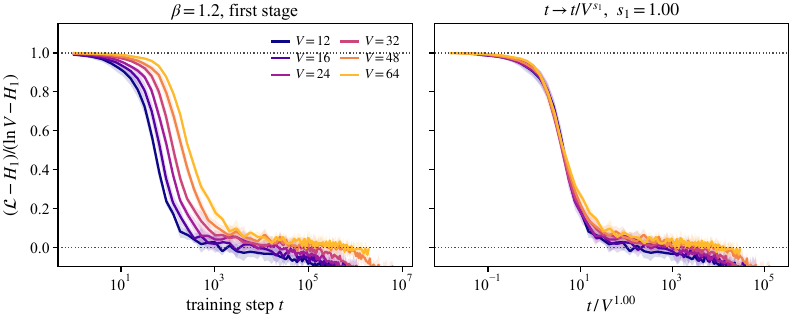}
\caption{\label{fig:app-b12-s1}\textbf{First stage of the dense grammar,
$\beta=1.2$.} Normalised loss between $\ln V$ and $H_1=H(X_4|X_3)$, for
$V=12$ to $64$ with $8$ realisations each, against the training step (left) and
against $t/V^{s_1}$ with $s_1=1.00\pm0.02$ (right). The collapse identifies the
first time scale as linear in the vocabulary size.}
\end{figure}

\begin{figure}[h!]\centering
\includegraphics[width=\textwidth]{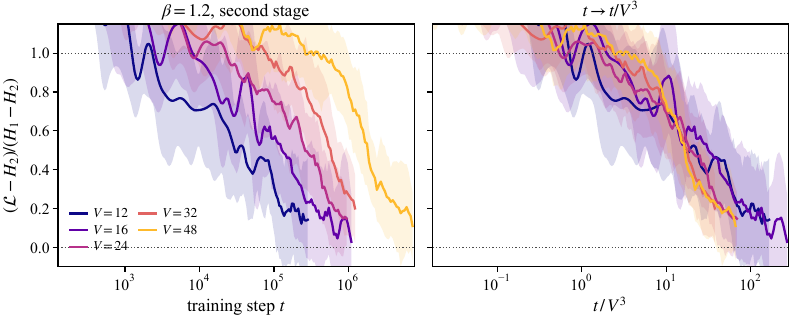}
\caption{\label{fig:app-b12-s2}\textbf{Second stage of the dense grammar,
$\beta=1.2$.} Normalised loss between $H_1$ and $H_2=H(X_4|X_1,X_2,X_3)$, for
$V=12$ to $48$ with $8$ realisations each, against the training step (left) and
against $t/V^{s_2}$ with $s_2=3.04\pm0.34$ (right). The curves sit at the
plateau for $t\ll V^{3}$ and reach the floor only for $t\gg V^{3}$, two powers
of $V$ later than the first stage. $V=64$ is omitted: it does not complete this
stage within the available budget. Curves are smoothed with a $5$-point moving
average in $\ln t$, the second stage being the noisier of the two.}
\end{figure}

\begin{figure}[h!]\centering
\includegraphics[width=\textwidth]{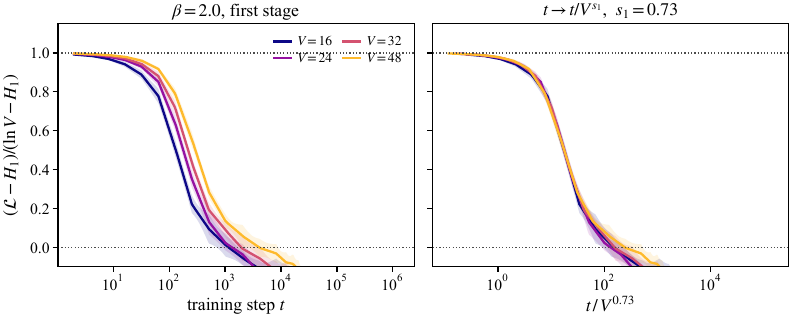}
\caption{\label{fig:app-b20-s1}\textbf{First stage,
$\beta=2$.} Same quantities as Fig.~\ref{fig:app-b12-s1}, for $V=16$ to $48$,
with $s_1=0.73\pm0.04$. These runs use the smaller model of
App.~\ref{app:protocol} and are not to be compared on the same axes with the
$\beta=1.2$ and $\beta=6$ curves.}
\end{figure}

\begin{figure}[h!]\centering
\includegraphics[width=\textwidth]{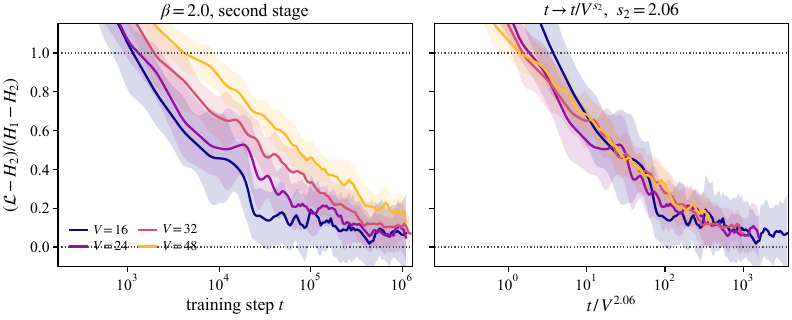}
\caption{\label{fig:app-b20-s2}\textbf{Second stage,
$\beta=2$.} Same quantities as Fig.~\ref{fig:app-b12-s2}, with
$s_2=2.06\pm0.37$. At $V=48$ the stage only partly completes within the
available budget, so the fit relies on the half-way crossing rather than on the
approach to the floor, and the uncertainty on $s_2$ is correspondingly larger.
At $\beta=2$ the two stages are separated by roughly $1.3$ powers of $V$,
against $2$ powers at $\beta=1.2$: the schedule is already contracting.}
\end{figure}

\begin{figure}[H]\centering
\includegraphics[width=\textwidth]{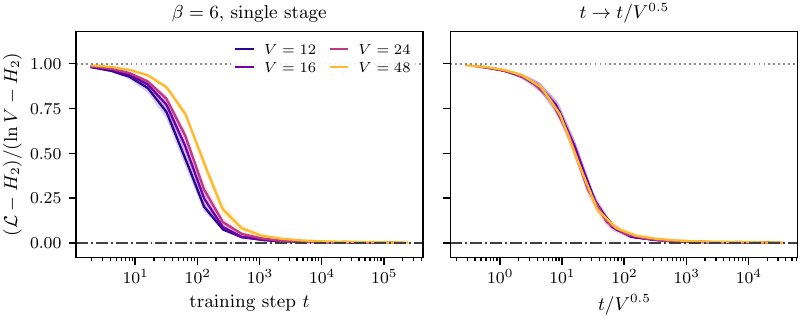}
\caption{\label{fig:app-b60}\textbf{Single stage of the sparse grammar,
$\beta=6$.} Normalised loss between $\ln V$ and $H_2$, for $V=12$ to $48$ with
$8$ realisations each, against the training step (left) and against $t/V^{s}$
with $s=0.47$ (right); fitting the single-stage curves directly gives
$s=0.44\pm0.03$. No plateau at $H_1$ is visible: the sibling and the cousins are
acquired together, and one exponent describes the whole descent.}
\end{figure}

\section{Adam time scales for a linear model at $L=1$}
\label{app:adam}

The exponents $1/2$ and $1$ of the first time scale (Fig.~\ref{fig:L1}) can be
obtained from an argument about the escape from initialisation of a linear
model trained with Adam. We consider the $L=1$ task of predicting $X_2$ from
$X_1$, for which the linear model is essentially the embedding layer of the
transformer; the same computation applies to the first stage at $L=2$ with
$(X_3,X_4)$ in place of $(X_1,X_2)$. The argument concerns the beginning of the
descent, not the approach to the entropy floor.

\paragraph{Linear model.}
The model is a table of logits $\theta\in\mathbb R^{V\times V}$, with
$\hat p_\theta(b\mid a)=e^{\theta_{ab}}/\sum_{b'}e^{\theta_{ab'}}$, initialised at
$\theta=0$, i.e. at the uniform predictor with loss $\ln V$. For one sample
$(X_1,X_2)$ the cross-entropy $\ell(\theta)=-\ln\hat p_\theta(X_2\mid X_1)$ has
gradient
\begin{equation}\label{eq:app-grad}
g_{ab}\coloneqq\partial_{\theta_{ab}}\ell(\theta)
=\mathbf 1\{X_1=a\}\big(\hat p_\theta(b\mid X_1)-\mathbf 1\{X_2=b\}\big).
\end{equation}
A batch of $B$ independent samples gives the minibatch gradient
$\bar g_{ab}=\frac1B\sum_{i=1}^B g^{(i)}_{ab}$, with mean and variance
\begin{equation}\label{eq:app-moments}
\mathbb E\,\bar g_{ab}=\mu_{ab}\coloneqq\mathbb E\,g_{ab},\qquad
\operatorname{Var}\bar g_{ab}=\frac{\sigma_{ab}^2}{B},\qquad
\sigma_{ab}^2\coloneqq\operatorname{Var}g_{ab}=\mathbb E\,g_{ab}^2-\mu_{ab}^2 .
\end{equation}

\paragraph{Optimisation model.}
Adam updates each coordinate by $-\eta\,m_{ab}/\sqrt{v_{ab}}$, where $m$ and
$v$ are exponential moving averages of $\bar g$ and $\bar g^{\,2}$. We replace
the running averages by population moments, which is justified while the
parameters move little over the averaging window and batches are fresh, as in
our protocol (App.~\ref{app:protocol}). The update then reads
\begin{equation}\label{eq:app-adam}
\Delta\theta_{ab}=-\eta\,\frac{\mu_{ab}}{\sqrt{\mu_{ab}^2+\sigma_{ab}^2/B}}
=-\eta\,\frac{\mathrm{SNR}_{ab}}{\sqrt{1+\mathrm{SNR}_{ab}^2}},\qquad
\mathrm{SNR}_{ab}\coloneqq\frac{\mu_{ab}\sqrt B}{\sigma_{ab}} .
\end{equation}
In the noise-dominated regime $|\mathrm{SNR}_{ab}|\ll1$, which holds for a
sufficiently small batch, the coordinate drifts at speed $\eta\,|\mathrm{SNR}_{ab}|$
per step. If it must move by a logit distance $D_{ab}$ to reach its target, the
number of steps is
\begin{equation}\label{eq:app-tstar}
t^*\asymp\frac{D_{ab}}{\eta}\,\frac{1}{|\mathrm{SNR}_{ab}|}
=\frac{D_{ab}}{\eta}\,\frac{\sigma_{ab}}{|\mu_{ab}|\sqrt B}.
\end{equation}
Plain SGD would drift at speed $\eta|\mu_{ab}|$ and take
$D_{ab}/(\eta|\mu_{ab}|)$ steps: the only difference is Adam's normalisation by
the root-mean-square gradient, which makes the speed depend on the ratio
$\mu/\sigma$ rather than on $\mu$. What follows is a computation of $\mu_{ab}$,
$\sigma_{ab}$ and $D_{ab}$ at $\theta=0$ in the two phases of the grammar.

\paragraph{Sparse phase.}
For $\beta\to\infty$ each parent has a single rule. Let the symbol $a$ be the
left child of $k$ parents, so that $\mathbb P(X_1=a)=k/V$; with high
probability the $k$ continuations are distinct, so
$\mathbb P(X_2=b\mid X_1=a)=1/k$ for the $k$ active $b$ and $0$ otherwise. At
$\theta=0$, $\hat p_\theta=1/V$ and Eq.~\eqref{eq:app-grad} gives, for an
active pair,
\begin{equation}\label{eq:app-sparse-moments}
\mu_{ab}=\frac kV\Big(\frac1V-\frac1k\Big)\simeq-\frac1V,\qquad
\mathbb E\,g_{ab}^2=\frac kV\Big[\frac1k\Big(1-\frac1V\Big)^2+\Big(1-\frac1k\Big)\frac1{V^2}\Big]\simeq\frac1V,
\end{equation}
so that $\sigma_{ab}^2\simeq1/V$ and
\begin{equation}\label{eq:app-sparse-snr}
|\mathrm{SNR}_{ab}|\simeq\sqrt{\frac BV}.
\end{equation}
The gradient is non-zero only on the fraction $1/V$ of samples with $X_1=a$,
where it is of order one: the signal is diluted by $1/V$ but the noise only by
$1/\sqrt V$. The inactive coordinates of the same row have $\mu_{ab'}=k/V^2$
and $\sigma_{ab'}^2\simeq k/V^3$, hence $|\mathrm{SNR}|\simeq\sqrt{kB/V}$, of
the same order, and drift downwards at the same speed. The target logits put
probability $1/k$ on the active continuations and $0$ on the others: the loss
of row $a$ falls from $\ln V$ to $\ln k$ when the gap between active and
inactive logits is of order $\ln V$, and reaches the half-way point used to
define $t^*$ at a gap of order $\tfrac12\ln V$. Hence $D\asymp\ln V$ and, from
Eq.~\eqref{eq:app-tstar},
\begin{equation}\label{eq:app-sparse-t}
t^*\asymp\frac{\ln V}{\eta}\sqrt{\frac VB}\;\propto\;V^{1/2}
\end{equation}
up to the logarithm, which is the exponent of the sparse regime.

\paragraph{Dense phase.}
For finite $\beta$ the joint distribution of the two tokens is
$\mathcal P^{(1)}_{ab}=\frac1V\sum_{z}\mathcal Q_z(a,b)$. Let
$C_{ab}\coloneqq\mathcal P^{(1)}_{ab}-\mathbb P(X_1=a)\,\mathbb P(X_2=b)$ be its
connected part and $\bar C\coloneqq\|C\|_F/V$ its root-mean-square entry. We call the phase dense when the conditional
distributions deviate little from uniform, i.e. when the typical relative
deviation
\begin{equation}\label{eq:app-eps}
\epsilon\coloneqq V\,\big|\mathbb P(X_2=b\mid X_1=a)-\tfrac1V\big|_{\rm typ}\simeq V^2\bar C
\end{equation}
is at most of order one; with the exact correlations, $\epsilon\approx0.8$ at
$\beta=1$ and $\epsilon$ between $0.95$ and $1.55$ at $\beta=1.2$ for $V$
between $16$ and $128$, so the estimate below applies at the border of its
validity in the dense runs. At $\theta=0$, Eq.~\eqref{eq:app-grad} gives
\begin{equation}\label{eq:app-dense-mu}
\mu_{ab}=\frac{\mathbb P(X_1=a)}{V}-\mathcal P^{(1)}_{ab}\eqqcolon-\Delta_{ab},
\end{equation}
where $\Delta_{ab}$ is the deviation of the joint from its row-uniform value,
equal to $C_{ab}$ up to the non-uniformity of the marginals, of typical size
$\bar C$. The uniform part of $\mathcal P^{(1)}_{ab}$ cancels because $\theta=0$
already is the uniform predictor: the only signal is the correlation. The
second moment is
\begin{equation}\label{eq:app-dense-sigma}
\mathbb E\,g_{ab}^2=\mathcal P^{(1)}_{ab}\Big(1-\frac1V\Big)^2
+\frac{\mathbb P(X_1=a)-\mathcal P^{(1)}_{ab}}{V^2}\simeq\mathcal P^{(1)}_{ab}\simeq\frac1{V^2},
\end{equation}
so $\sigma_{ab}\simeq1/V$: the noise is set by the joint probability itself,
not by its deviation. Therefore
\begin{equation}\label{eq:app-dense-snr}
|\mathrm{SNR}_{ab}|\simeq V\bar C\sqrt B .
\end{equation}
The target logits are $\theta^*_{ab}=\ln\mathbb P(X_2=b\mid X_1=a)+\mathrm{const}$,
so the coordinate must move by the relative deviation of the conditional from
uniform,
\begin{equation}\label{eq:app-dense-D}
D_{ab}=\big|\ln\big(V\,\mathbb P(X_2=b\mid X_1=a)\big)\big|
\simeq V\Big|\mathbb P(X_2=b\mid X_1=a)-\frac1V\Big|
=\frac{V|\Delta_{ab}|}{\mathbb P(X_1=a)}\simeq V^2\bar C .
\end{equation}
Signal and distance are both proportional to the correlation, which cancels in
Eq.~\eqref{eq:app-tstar}:
\begin{equation}\label{eq:app-dense-t}
t^*\asymp\frac{V^2\bar C}{\eta\,V\bar C\sqrt B}=\frac{V}{\eta\sqrt B}\;\propto\;V .
\end{equation}
The exponent is $1$ throughout the dense phase, whatever the value of $\bar C$
and hence of $\beta$. Under SGD, whose speed is $\eta|\mu_{ab}|\simeq\eta\bar C$,
the same distance takes $V^2/\eta$ steps.

\paragraph{Remarks.}
(i) The two results are the two limits of one formula. As $\beta$ grows at fixed
$V$ the relative deviations $\epsilon=V^2\bar C$ exceed one, the logit distance
saturates at $D=O(\ln V)$ while the signal-to-noise ratio $V\bar C\sqrt B$ keeps
growing, and Eq.~\eqref{eq:app-tstar} gives $t^*\propto V^{a_1-1}$ if
$\bar C\propto V^{-a_1}$; with $a_1=3/2$ for $\beta\ge2$ this is
Eq.~\eqref{eq:app-sparse-t}. The first-stage exponent therefore interpolates
from $1$ to $1/2$ as sparsity increases.
(ii) The noise-dominated regime assumed in Eq.~\eqref{eq:app-tstar} requires
$B\lesssim\sigma_{ab}^2/\mu_{ab}^2$, i.e. $B\lesssim V$ in the sparse phase and
$B\lesssim 1/(V\bar C)^2$ in the dense phase. Above these values, Adam reduces to
sign descent, $\Delta\theta_{ab}=-\eta\,\mathrm{sign}(\mu_{ab})$, and
$t^*\asymp D/\eta$ no longer grows as a power of $V$.

\begin{figure}[h!]
  \centering
  \includegraphics[width=\textwidth]{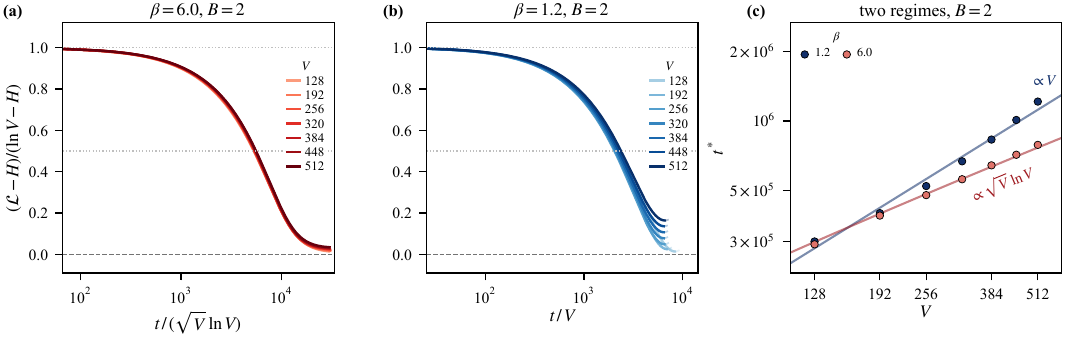}
  \caption{\label{fig:app-linear-B2}\textbf{Linear model at small batch size, $B=2$.}
  Normalised loss $(\mathcal L-H)/(\ln V-H)$ of the linear model, averaged over 8 grammars, for $V=128$ to $512$:
  (a) $\beta=6$ against $t/(\sqrt V\ln V)$, (b) $\beta=1.2$ against $t/V$. (c) Time $t^*$ at which the averaged
  loss first reaches $1/2$, with Eqs.~\eqref{eq:app-sparse-t} and~\eqref{eq:app-dense-t}, $t^*\propto\sqrt V\ln V$
  and $t^*\propto V$, prefactors fitted. Adam with its default parameters ($\beta_1=0.9$, $\beta_2=0.999$, $\epsilon=10^{-8}$, no weight decay; App.~\ref{app:protocol}), learning rate $\eta=3\times10^{-5}$, fresh samples at every step. At $\beta=1.2$ the loss levels off above the floor at late times and then rises (faint
  lines), more so at larger $V$; this does not affect $t^*$.}
\end{figure}

To test these scalings where the argument applies, Fig.~\ref{fig:app-linear-B2} shows the linear model itself
at batch size $B=2$, for which both conditions of remark (ii) hold over the whole range of $V$: $t^*$ grows as
$\sqrt V\ln V$ in the sparse phase and as $V$ in the dense one, as Eqs.~\eqref{eq:app-sparse-t}
and~\eqref{eq:app-dense-t} predict, and the descent of the loss collapses when time is divided accordingly.
The transformers of the previous sections are trained at larger batch sizes, where $B\lesssim V$ fails in the
sparse phase, and for them we use the argument only as a heuristic for the two exponents.

\section{Role of depth}
\label{app:depth}

The acquisition of a hierarchical grammar in stages suggests that a network
needs internal depth to represent the intermediate level of the tree. We test
this by training transformers with one and with two attention blocks on
\emph{identical} grammars: for every vocabulary size and every repetition, the
depth-$1$ and the depth-$2$ network see the same rule tensor, so the comparison
is paired and no difference can be attributed to the draw of the grammar. Both
panels below show the normalised loss $(\mathcal L-H_2)/(\ln V-H_2)$, with
$H_2=H(X_4|X_1,X_2,X_3)$ the full-context entropy of that realisation, averaged
over repetitions with bands giving one standard deviation.

\begin{figure}[h!]\centering
\includegraphics[width=\textwidth]{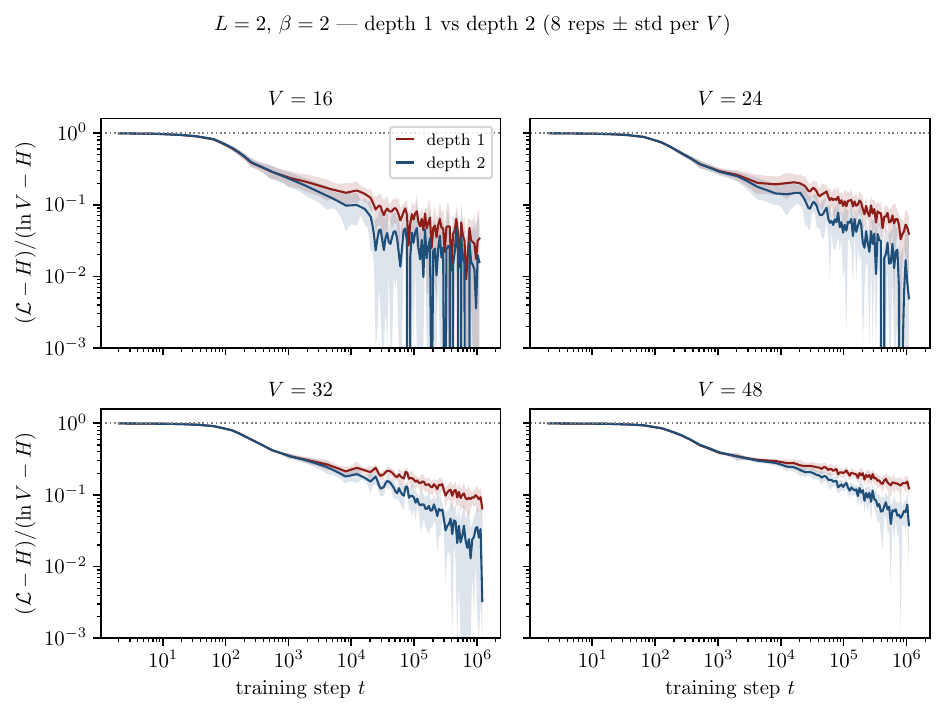}
\caption{\label{fig:app-depth-b20}\textbf{At the beta 2.0, depth helps.}
Depth-$1$ and depth-$2$ transformers trained on the same grammars at $\beta=2$,
for $V=16,24,32,48$ with $8$ repetitions each. The two networks are
indistinguishable while the loss falls from the uniform predictor, and separate
at long times: the depth-$2$ network reaches the full-context floor, whereas the
depth-$1$ network stalls above it. Smaller model of App.~\ref{app:protocol}
(embedding dimension $128$, batch size $64$).}
\end{figure}

\begin{figure}[!h]\centering
\includegraphics[width=\textwidth]{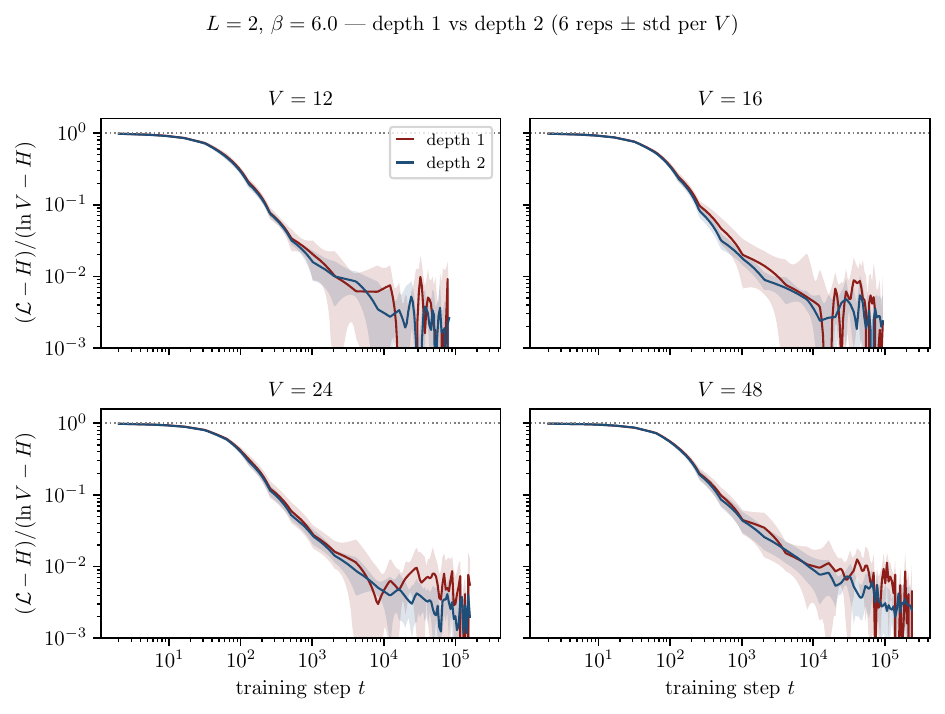}
\caption{\label{fig:app-depth-b60}\textbf{In the sparse regime, depth is
irrelevant.} The same comparison at $\beta=6$, for $V=12,16,24,48$ with $6$
paired repetitions each, on a logarithmic vertical scale so that the approach to
the floor is visible over three decades. The two depths are indistinguishable
throughout, the depth-$1$ network being at most marginally slower. Protocol of
the main text (embedding dimension $256$, batch size $128$); the depth-$1$ runs
were trained on the rule tensors of the depth-$2$ runs, read from file, so the
grammars agree exactly.}
\end{figure}

The contrast follows the picture of the main text. When the grammar is dense
enough for the levels to be acquired one at a time, the second stage requires
composing the two latent symbols, and a network with a single attention block
cannot represent that composition well enough to reach the Bayes floor. When the
grammar is sparse, the whole sequence is acquired in a single step and no
intermediate representation is needed, so the second block buys nothing. Depth
therefore pays off exactly in the regime in which learning is hierarchical in
time.

We note a limitation: no paired depth-$1$/depth-$2$ runs exist at $\beta=2$ in
the configuration of the main text, so Fig.~\ref{fig:app-depth-b20} uses the
smaller model while Fig.~\ref{fig:app-depth-b60} uses the main one. The
comparison between depths is paired and exact within each figure; the comparison
of the two figures with each other is not at fixed architecture.

%%%%%%%%%%%%%%%%%%%%%%%%%%%%%%%%%%%%%%%%%%%%%%%%%%%%%%%%%%%%

\newpage

\end{document}